\documentclass[sigconf]{acmart}
\AtBeginDocument{%
  }

\usepackage{cleveref}
\usepackage{multirow}
\definecolor{projectlink}{HTML}{FB3199}

\copyrightyear{2026}
\acmYear{2026}
\setcopyright{cc}
\setcctype{by-nc-nd}
\acmConference[MM '26]{Proceedings of the 34th ACM International Conference on Multimedia}{November 10--14, 2026}{Rio de Janeiro, Brazil}
\acmBooktitle{Proceedings of the 34th ACM International Conference on Multimedia (MM '26), November 10--14, 2026, Rio de Janeiro, Brazil}
\acmDOI{10.1145/3767308.3835899}
\acmISBN{979-8-4007-2213-4/2026/11}

\begin{document}

\title{VibeAvatar: Aligning Phonetic Kinematics and Human Aesthetics for High-Fidelity Talking Avatar Synthesis}


\author{Qilin Wang}
\authornote{These authors contributed equally to this work.}
\affiliation{%
  \institution{School of Computer Science, \\Peking University}
  \city{Beijing}
  \country{China}}
\email{qilinwang25@stu.pku.edu.cn}

\author{Mingyu Li}
\authornotemark[1]
\affiliation{%
  \institution{School of Electronics Engineering and Computer Science, Peking University}
  \city{Beijing}
  \country{China}}
\email{mingyulics@stu.pku.edu.cn}

\author{Hao Tang}
\authornote{Corresponding author.}
\affiliation{%
  \institution{School of Computer Science, \\Peking University}
  \city{Beijing}
  \country{China}}
\email{bjdxtanghao@gmail.com}

\makeatletter
\patchcmd{\@mkauthors@iii}
  {\addresses\let\and\@typeset@author@bx\and\par\bigskip}
  {\addresses\let\and\@typeset@author@bx\and\par\smallskip
   \noindent\href{https://kelu007.github.io/vibe-avatar/}{\textcolor{projectlink}{\nolinkurl{https://kelu007.github.io/vibe-avatar/}}}\par\bigskip}
  {}{\PackageError{sigconf_arxiv}{Could not place the project URL below the affiliations}{}}
\makeatother


\begin{abstract}

    Multi-modal talking avatar synthesis aims to generate realistic talking videos from a reference portrait and speech. Despite rapid progress in diffusion-based methods, existing approaches still struggle to jointly achieve accurate lip articulation, human-preferred motion aesthetics, and efficient inference. We observe that phonetic accuracy and motion aesthetics arise from fundamentally different sources and should be addressed at complementary stages rather than learned implicitly by a single generator. Based on this insight, we propose VibeAvatar, which disentangles these two objectives through a Phonetic Kinematics Adapter (PKA) that converts recognition-oriented speech features into phonetic-kinematic conditions at the conditioning stage, and an Aesthetic Motion Policy (AMP) that optimizes a flow-consistent stochastic sampling policy via Group Relative Policy Optimization (GRPO) at the post-training stage. With a lightweight flow-based motion generator operating in a compact 1D warp-based latent motion space, VibeAvatar achieves state-of-the-art results in articulation, aesthetics, and efficiency on both objective metrics and user studies, while generating a 10-second 512\,px video in under 10 seconds with only $\sim$3\,GB VRAM.
    
\end{abstract}


\begin{teaserfigure}
\centering
  \includegraphics[width=0.95\textwidth]{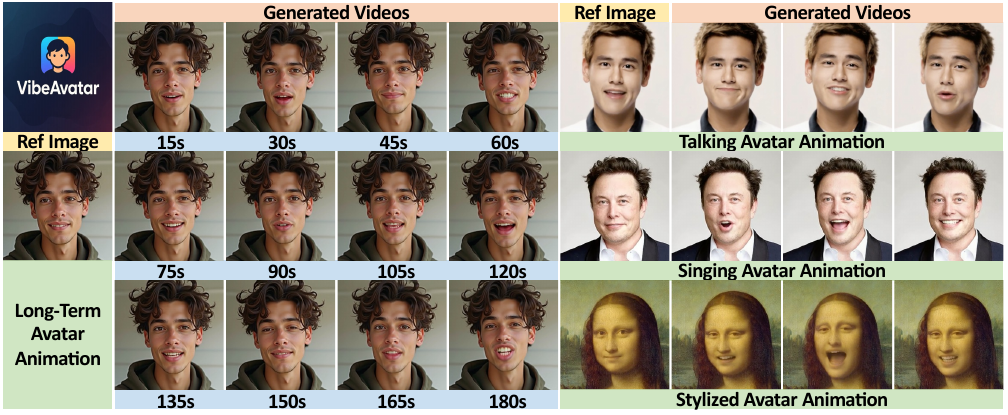}
  \caption{VibeAvatar synthesizes high-fidelity talking avatars with accurate lip articulation, human-preferred motion aesthetics, and efficient inference. It preserves stable identity and natural facial dynamics across speech and singing scenarios.}
  \label{fig:teaser}
\end{teaserfigure}


\maketitle

\section{Introduction}
\label{sec:intro}

Talking avatar synthesis is a multi-modal generation task that produces a talking human video by jointly reasoning over a reference portrait and a speech signal. It serves as a core technology for digital humans in entertainment, education, and virtual interaction. A practical system should simultaneously achieve accurate lip articulation, human-aligned motion aesthetics, and efficient inference.

Driven by rapid advances in diffusion-based generative frameworks~\cite{dhariwal2021diffusion,ddpm,ddim,nichol2021improved,song2020score}, talking avatar synthesis has progressed substantially. Early studies~\cite{dreamtalk,aniportrait,sadtalker,videoretalking,toontalker,stylesync,dpe,vexpress,styleheat} typically relied on two-stage pipelines that first predicted facial motion parameters, such as 3D Morphable Models~\cite{3dmm} (3DMM) coefficients, followed by a rendering stage. While interpretable, these cascaded frameworks often produced rigid and emotionless facial movements. More recent end-to-end diffusion models~\cite{echomimic,hallo3,loopy,emo,sonic,latentsync,cyberhost,fantasytalking,wan-s2v} learn direct audio--visual correlations and further close the gap between synthesized and real talking videos.

\begin{figure}[tbp]
    \centering
    \includegraphics[width=0.95\linewidth]{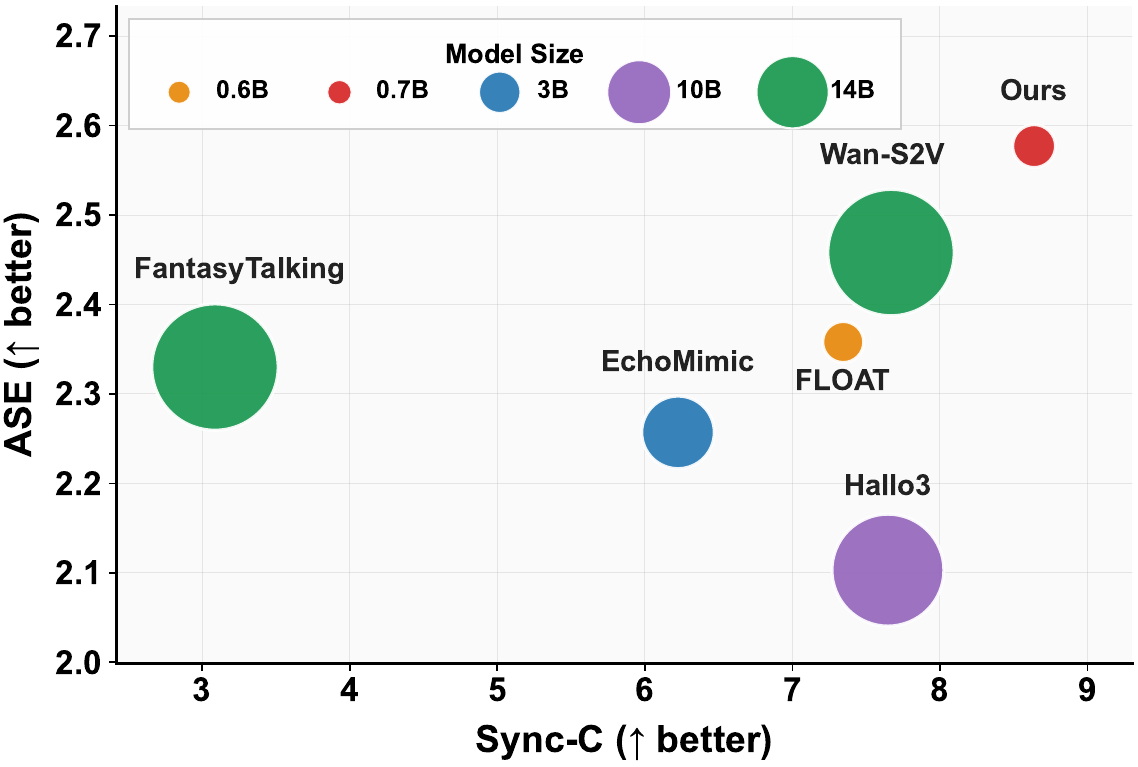}
    \caption{
        Comparison of recent talking avatar methods on ASE~\cite{qalign} and Sync-C~\cite{sync}. Existing methods still struggle to jointly improve aesthetic quality and lip synchronization.
    }
    \label{fig:intro_comparison}
    \vspace{-0.2cm}
\end{figure}

However, these end-to-end models still face three unresolved issues. (i) Generic speech encoders are primarily optimized for recognition semantics rather than facial kinematics, so the generator must recover phonetic timing and co-articulation implicitly, making precise lip articulation difficult. (ii) Articulation accuracy alone does not guarantee human-preferred motion: viewers also judge global expressiveness, temporal rhythm, and facial dynamics, yet standard training objectives provide no explicit signal to optimize these perceptual factors. (iii) Many recent models operate in high-dimensional pixel or latent-video space and rely on iterative spatial-temporal denoising, tying quality gains to heavier backbones and more sampling steps at the cost of practical efficiency. As shown in \Cref{fig:intro_comparison}, recent methods~\cite{fantasytalking,echomimic,float,wan-s2v,hallo3} still reveal a clear trade-off between synchronization and aesthetic quality.

These observations suggest that the bottleneck is not simply model capacity. Fine-grained lip articulation is largely governed by phonetic structure and co-articulation, whereas motion aesthetics is a higher-level perceptual property shaped by global dynamics and human preference. Their optimal solutions therefore arise from different sources, and asking a single generator to learn both implicitly makes the trade-off hard to resolve. The key insight behind VibeAvatar is to treat phonetic accuracy and aesthetic motion quality as two related but distinct problems: the former should be handled at the conditioning stage by adapting speech representations to facial kinematics, while the latter should be addressed by post-training the sampling policy with human preference alignment. Once these two factors are disentangled, neither requires the generator itself to be large: a lightweight motion backbone suffices because phonetic precision is supplied by the adapted conditioning and aesthetic quality is injected through preference-driven policy optimization, rather than being learned implicitly through model capacity.

Based on this insight, we propose VibeAvatar, an efficient framework for multi-modal talking avatar synthesis that jointly improves lip articulation, motion aesthetics, and inference efficiency. Our method is built upon three coordinated designs: (i) a \textbf{Phonetic Kinematics Adapter (PKA)} that uses hierarchical shifted attention and a learnable kinematic gate to convert generic speech features into phonetic-kinematic conditions, bridging the gap between recognition-oriented representations and articulatory motion; (ii) an \textbf{Aesthetic Motion Policy (AMP)} that reformulates flow-based sampling as a stochastic policy and optimizes it via Group Relative Policy Optimization~\cite{grpo} (GRPO) within a flow-consistent neighborhood, with a timestep-truncated strategy that restricts preference optimization to early coarse-grained steps so that fine-grained lip details are preserved; and (iii) a lightweight \textbf{flow-based motion generator} operating in a compact 1D warp-based latent space, paired with a two-stage training strategy that first establishes a stable articulation-aware prior and then aligns motion dynamics with human aesthetics without degrading phonetic precision.

With these designs working together, VibeAvatar establishes stronger performance on lip articulation, motion aesthetics, and efficiency, as shown in \Cref{fig:teaser}. Extensive experiments further demonstrate that our method achieves better qualitative and quantitative results.
In summary, our contributions are as follows:

\begin{itemize}

    \item We present VibeAvatar, which disentangles phonetic conditioning from aesthetic alignment, yielding a favorable quality--efficiency balance within a lightweight framework.
    
    \item We propose a Phonetic Kinematics Adapter (PKA) with hierarchical shifted attention and a learnable kinematic gate to convert speech features into phonetic-kinematic conditions for accurate lip articulation.
    
    \item We propose an Aesthetic Motion Policy (AMP) that optimizes a flow-consistent stochastic sampling policy via GRPO with a timestep-truncated strategy, aligning motion with human aesthetics without degrading articulation.

    \item Experiments show state-of-the-art results on both objective metrics and user studies, with under-10-second inference for a 10-second 512\,px video and $\sim$3\,GB VRAM.
    
\end{itemize}

\section{Related Work}
\label{sec:related_work}

\begin{figure*}[ht]
    \centering
    \includegraphics[width=0.95\linewidth]{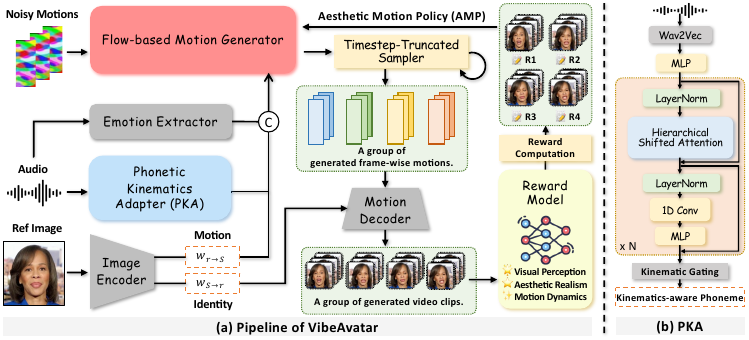}
    \vspace{-0.2cm}
    \caption{
    Overview of VibeAvatar. (a) The framework integrates a lightweight flow-based motion generator, a Phonetic Kinematics Adapter (PKA), and an Aesthetic Motion Policy (AMP) to improve lip articulation, motion aesthetics, and inference efficiency.
    (b) PKA transforms generic speech features into phonetic-kinematic conditions through hierarchical shifted attention, enabling local bidirectional interaction, causal aggregation, and a progressively enlarged receptive field for accurate lip articulation.
    }
    \label{fig:framework}
    \vspace{-0.2cm}
\end{figure*}

\noindent\textbf{Controllable Video Generation.}
Recent advances in controllable video generation have been fueled by diffusion-based image synthesis models~\cite{dhariwal2021diffusion,esser2024scaling,SD,hertz2022prompt,tumanyan2023plug}. Early works~\cite{VDM,brooks2024video,harvey2022flexible,makeavideo,motioneditor,tuneavideo} extended pretrained UNet-based diffusion backbones into the temporal domain by adding motion-aware or temporal attention layers for joint spatiotemporal modeling. Representative methods such as Stable Video Diffusion~\cite{SVD}, Make-A-Video~\cite{makeavideo}, and AnimateDiff~\cite{animatediff} achieved high-quality short video generation by coupling spatial priors with temporal modeling. More recent efforts transition toward transformer-based diffusion frameworks for improved scalability and temporal reasoning. The Diffusion-in-Transformer (DiT) family~\cite{hunyuanvideo,wan,easyanimate,cogvideox,vidu,goku} replaces convolutional UNets with transformer blocks, exhibiting stronger modeling capacity for complex video dynamics. Large-scale frameworks such as CogVideoX~\cite{cogvideox}, HunyuanVideo~\cite{hunyuanvideo}, and Wan~\cite{wan} further integrate multimodal conditions such as text, audio, and pose within hierarchical or dual-stream pipelines, enhancing controllability across tasks including identity preservation~\cite{yuan2025identity}, expression control~\cite{ma2024follow}, and virtual try-on~\cite{li2025pursuing}. While these methods provide powerful generative backbones, they mainly target general video realism and controllability rather than the task-specific balance among lip articulation, motion aesthetics, and efficiency required by talking avatars.

\noindent\textbf{Audio-driven Avatar Animation.} 
Audio-driven avatar animation aims to generate realistic talking human videos from speech. Early studies~\cite{dreamtalk,aniportrait,sadtalker,videoretalking,toontalker,stylesync,dpe,vexpress,styleheat，consistentavatar} employed two-stage pipelines that predicted facial motion parameters based on 3D Morphable Models~\cite{3dmm} (3DMM) or landmarks and then rendered frames. Although interpretable, such cascaded methods exhibited limited expressiveness, often resulting in rigid lip movements and insufficient emotional dynamics.
Recent advances have been driven by diffusion-based end-to-end frameworks~\cite{echomimic,hallo3,loopy,emo,sonic,latentsync,cyberhost,fantasytalking,wan-s2v} that directly learn audio--visual correlations for coherent speech-to-video generation. Among them, FLOAT~\cite{float} enhances audio-lip synchronization and temporal coherence through multi-stage refinement. Hallo3~\cite{hallo3} introduces a diffusion transformer architecture for highly dynamic and photorealistic portrait animation. EchoMimic~\cite{echomimic} enables editable landmark-conditioned synthesis for fine-grained local control. FantasyTalking~\cite{fantasytalking} extends the task toward coherent facial and upper-body motion with improved identity preservation, while Wan-S2V~\cite{wan-s2v} demonstrates the strong generative potential of large commercial video models in portrait animation.
Despite recent advances, existing methods still struggle to jointly achieve accurate lip articulation, human-aligned motion aesthetics, and efficient inference in talking avatar synthesis. In contrast, our work explicitly separates phonetic conditioning from aesthetic alignment, enabling a lightweight generator to focus on stable motion prediction while dedicated designs improve lip articulation and human-preferred motion dynamics.

\section{The Proposed Method}
\label{sec:method}

\subsection{Overview}

Given a reference image and an audio clip, VibeAvatar synthesizes talking-avatar videos with accurate lip articulation, human-aligned motion aesthetics, and efficient inference. 
As shown in \Cref{fig:framework}, our framework achieves these goals through three coordinated designs. A flow-based motion generator (\Cref{sec:model_arch}) predicts motion efficiently in a compact latent space. A Phonetic Kinematics Adapter (PKA, \Cref{sec:audio_adapter}) transforms raw speech features into motion-oriented conditions that preserve phonetic details for lip articulation. An Aesthetic Motion Policy (AMP, \Cref{sec:stochastic_policy}) reformulates flow-based sampling as a stochastic policy and optimizes it via GRPO~\cite{grpo} toward human-preferred dynamics. Finally, the two-stage training strategy (\Cref{subsec:training_strategy}) first learns a motion prior and then aligns motion sampling with human aesthetic preference.

\subsection{Flow-based Motion Generator}
\label{sec:model_arch}

As shown in \Cref{fig:framework}(a), our flow-based motion generator disentangles identity from facial dynamics and predicts motion in a compact latent space. It applies conditioning at the frame level for precise motion control. These designs mainly provide a stable and efficient foundation for the later PKA and AMP modules.

\noindent\textbf{1D Motion Representation.}
To preserve identity while keeping motion generation lightweight, we represent facial dynamics in the latent warp space of LIA~\cite{lia}. Given a reference image $S\!\in\!\mathbb{R}^{3\times H\times W}$, the autoencoder produces an identity warp, a motion warp, and multi-scale features:
\begin{equation}
\text{Enc}:\; S \mapsto \bigl(w_{S\to r},\, w_{r\to S},\, F_S\bigr).
\end{equation}
Here $w_{S\to r}\!\in\!\mathbb{R}^{d_w}$ encodes identity as the warp from $S$ to a canonical neutral pose $r$, $w_{r\to S}\!\in\!\mathbb{R}^{d_w}$ encodes motion as the warp from $r$ back to $S$, and $F_S$ preserves high-frequency appearance details. We further expand the motion code on a learnable, low-dimensional orthogonal basis $\Lambda=\{\lambda_1,\ldots,\lambda_m\}$, yielding a compact \emph{1D} trajectory for facial dynamics. To animate $S$ with a driver $D$, we decode
\begin{equation}
    \text{Dec}:\; \bigl(w_{S\to r},\, w_{r\to D},\, F_S\bigr)\mapsto \hat{D}(S),
\end{equation}
which preserves identity via $(w_{S\to r},F_S)$ while substituting motion via $w_{r\to D}$. This factorization confines controllable facial dynamics to a low-dimensional space, which reduces computational cost and makes the backbone well suited for efficient inference.

\noindent\textbf{Flow-based Motion Generation.}
To generate coherent facial motion with low overhead, we predict motion latents of $L$ consecutive frames jointly rather than reconstructing each frame independently. Let the framewise audio conditions be $c\!\in\!\mathbb{R}^{L\times d_c}$, the past $L'$ motion latents be $W_{\mathrm{past}}\!\in\!\mathbb{R}^{L'\times d_w}$, and the reference motion warp of image $S$ be $w_{r\to S}\!\in\!\mathbb{R}^{d_w}$. We form a trajectory $z_t\!\in\!\mathbb{R}^{(L'+L)\times d_w}$ and learn a conditional vector field
$
   v_\theta\bigl(z_t,\, t;\, W_{\mathrm{past}},\, w_{r\to S},\, c\bigr), 
$
solving the ODE
\begin{equation}
\frac{d z_t}{dt} = v_\theta\bigl(z_t,\, t;\, W_{\mathrm{past}},\, w_{r\to S},\, c\bigr), 
\qquad z_{1}\sim\mathcal{N}(0,I),
\end{equation}
to obtain $z_{0}$. The last $L$ rows of $z_{0}$ are the predicted motion latents for the current clip, while the leading $L'$ rows provide temporal context to enforce a smooth transition. This clip-wise latent prediction keeps the backbone compact while supplying stable motion trajectories for the later PKA conditioning and AMP optimization.

\noindent\textbf{Framewise Conditioning via AdaLN.}
We implement the flow-based motion generator as a framewise DiT~\cite{dit}, which is well suited for modeling latent motion tokens. Conditioning is then injected into each block through adaptive LayerNorm. For the $i$-th frame in the $j$-th DiT block, we predict adaptive LayerNorm parameters $(\alpha_i^{(j)},\beta_i^{(j)},\gamma_i^{(j)})$ from a linear head fed by the fused conditioning. The scale--shift pair $(\alpha,\beta)$ affords fine-grained frame-level control, while the gate $\gamma$ modulates the residual pathway to stabilize conditioning strength. Each block also includes localized inter-frame cross-attention over a small temporal window, ensuring temporal coherence without sacrificing per-frame precision.

\noindent\textbf{Emotion Extractor.}
To complement phonetic conditioning with expression cues, we additionally incorporate an emotion prior from the audio stream. Specifically, we apply the pretrained classifier of~\cite{emotion} to the frame-aligned audio features, embed the predicted emotion label, and concatenate it with the other conditioning signals. This simple design supplies prosody-related expression information to DiT without introducing a heavy auxiliary branch.

\subsection{Phonetic Kinematics Adapter (PKA)}
\label{sec:audio_adapter}

Accurate lip articulation requires audio conditions that preserve fine-grained co-articulation while respecting utterance-level temporal causality. However, self-supervised encoders such as Wav2Vec~\cite{wav2vec} produce recognition-oriented frame-wise representations whose semantics are indirectly related to facial kinematics. Directly conditioning the motion backbone on these raw features forces it to recover phonetic timing, cross-phoneme transitions, and motion saliency by itself, leading to imprecise or unstable mouth motion.

\begin{figure}[!t]
    \centering
    \includegraphics[width=\linewidth]{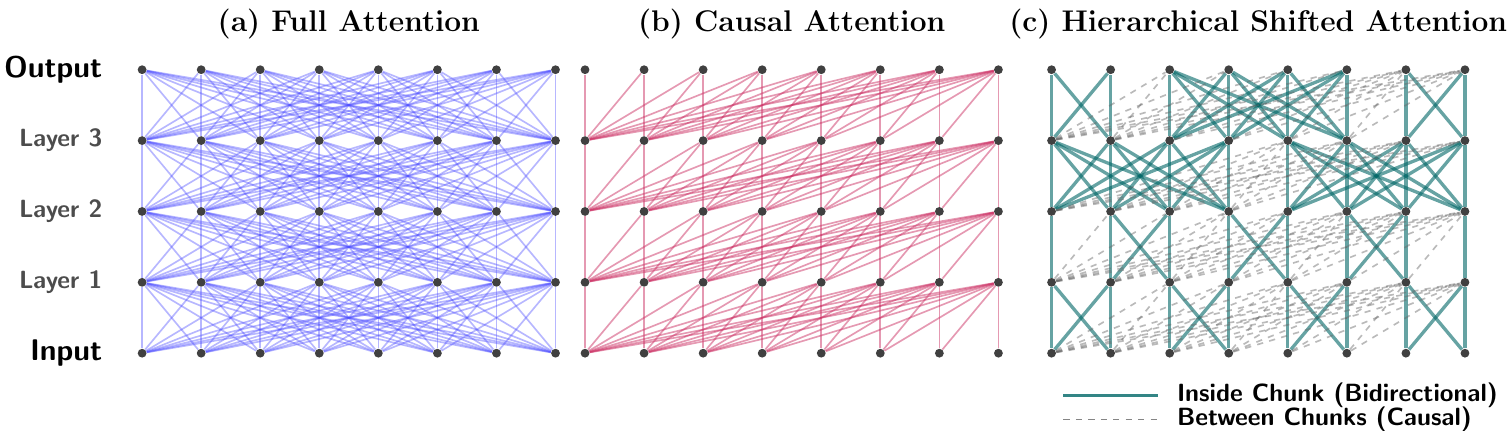}
    \caption{
    Comparison of audio context modeling schemes. (a) Full Attention captures global context but incurs high cost and leaks future information. (b) Causal Attention preserves causality but weakens local bidirectional interaction. (c) Our Hierarchical Shifted Attention supports bidirectional interaction within chunks and causal aggregation across chunks, while progressively enlarging the receptive field.
    }
    \label{fig:audio_modeling}
    \vspace{-0.4cm}
\end{figure}

The Phonetic Kinematics Adapter (PKA) resolves this mismatch by explicitly adapting generic speech representations into phonetic-kinematic conditions before they are injected into the flow-based backbone. Full attention in ~\Cref{fig:audio_modeling}(a) enriches context but leaks future information, while causal attention in ~\Cref{fig:audio_modeling}(b) preserves causality but weakens local bidirectional interactions that are important for co-articulation. To balance these two extremes, PKA introduces two key designs. First, \emph{Hierarchical Shifted Attention} progressively enlarges chunk size across layers and applies alternating shifted windows, enabling bidirectional interaction inside each chunk and causal aggregation between chunks, as shown in ~\Cref{fig:audio_modeling}(c). Second, a \emph{Learnable Kinematic Gate} filters motion-irrelevant semantics and amplifies articulation-related variations. Together, these designs convert raw speech features into phonetic-kinematic conditions that are more suitable for lip articulation.

\noindent\textbf{Hierarchical Shifted Attention.}
As illustrated in \Cref{fig:framework}(b), the raw frame-wise audio features $A_{\mathrm{raw}} \in \mathbb{R}^{L \times d_{\mathrm{audio}}}$ are first projected into the adapter space as $A^{(0)} \in \mathbb{R}^{L \times d}$. PKA then applies $N$ stacked attention blocks with progressively enlarged chunk sizes. In the $l$-th block, the chunk size grows as
\begin{equation}
K^{(l)} = K_{\mathrm{base}} \times 2^{\lfloor (l-1)/2 \rfloor},
\end{equation}
and the temporal shift is set to $S^{(l)} = 0$ for odd-numbered blocks and $S^{(l)} = K^{(l)}/2$ for even-numbered. This hierarchical schedule expands the receptive field from local phoneme transitions to broader speech context. And the shifted window mechanism ensures that phonetic events divided by a chunk boundary in layer $l-1$ are centered within the chunk of layer $l$, reducing boundary fragmentation for phonetic events that straddle chunk borders.

Given the shifted partition, we construct the chunk-causal attention mask $M^{(l)}$ that determines the visibility between a query $i$ and a key $j$ as:
\begin{equation}
M^{(l)}_{i,j} =
\begin{cases}
0, & \text{if } \left\lfloor \frac{j + S^{(l)}}{K^{(l)}} \right\rfloor \le \left\lfloor \frac{i + S^{(l)}}{K^{(l)}} \right\rfloor,\\
-\infty, & \text{otherwise}.
\end{cases}
\end{equation}
The forward pass of the $l$-th block is then written as
\begin{align}
H^{(l)} &= A^{(l-1)} + \operatorname{Attn}\!\left(\operatorname{LN}\!\left(A^{(l-1)}\right);\, M^{(l)}\right),\\
A^{(l)} &= H^{(l)} + \operatorname{MLP}\!\left(
    \operatorname{Conv1D}\!\left(\operatorname{LN}\!\left(H^{(l)}\right)\right)
\right).
\end{align}
This design preserves bidirectional interactions within each local chunk, enforces causal aggregation across chunks, and expands context hierarchically from fine co-articulation to broader prosodic structure. The lightweight Conv1D+MLP branch further smooths temporal features and improves local mixing, making the resulting representation both articulation-aware and causally consistent.

\noindent\textbf{Learnable Kinematic Gate.}
After hierarchical refinement, we further adapt the audio representation with a learnable kinematic gate. Instead of passing all semantic content directly to the motion backbone, the gate predicts frame-wise scale and shift factors $(\alpha_t, \beta_t)$ from the refined features and modulates them as $\tilde{c}_t = \alpha_t \odot A^{(N)}_t + \beta_t$. This modulation emphasizes articulation-related transitions and de-emphasizes silent or weakly informative regions. The zero-initialized gate provides a stable conditioning path at the start of training and gradually learns to transform generic speech representations into motion-driving cues. The final PKA output is used as the conditioning signal $c$ for the flow-based motion generator.

\subsection{Aesthetic Motion Policy (AMP)}
\label{sec:stochastic_policy}

A remaining challenge is that a standard flow-matching sampler is almost deterministic once the condition and initial noise are fixed. Such low-entropy sampling provides little room to explore alternative motion realizations, making it difficult to align facial dynamics with human aesthetic preference. To address this issue, we introduce an Aesthetic Motion Policy (AMP), which reformulates flow-based sampling as a stochastic policy while preserving consistency with the learned motion field. AMP enables preference optimization over motion trajectories without abandoning the strong articulation prior learned by the motion generator and PKA.

\noindent\textbf{Flow-consistent Stochastic Motion Sampling.}
Let $z_1$ denote the initial noise latent at pseudo-time $t=1$, $z_0$ the terminal latent at $t=0$, and $z_t$ the latent at intermediate time $t \in [0,1]$. The base sampler uses a deterministic Euler update
\begin{equation}
z_{\alpha}
= z_t - v_\theta\bigl(z_t, t; W_{\mathrm{past}}, w_{r\to S}, c\bigr)\,(t-\alpha),
\label{eq:deterministic_euler}
\end{equation}
which produces same trajectories given fixed noise and conditions.

AMP introduces controlled exploration by replacing this deterministic transition with a flow-consistent stochastic update. We construct two extrapolated endpoints:
\begin{align}
\hat{z}_0 &= z_t - t\, v_\theta\bigl(z_t, t; W_{\mathrm{past}}, w_{r\to S}, c\bigr),\\
\hat{z}_1 &= z_t + (1-t)\, v_\theta\bigl(z_t, t; W_{\mathrm{past}}, w_{r\to S}, c\bigr),
\end{align}
which approximate the latent states toward $t=0$ and $t=1$, respectively. For the backward step $t \to \alpha = t-\Delta t$, we then sample
\begin{equation}
z_{\alpha}
= \mu_\theta(z_t, t, \alpha)
+ \sigma(t)\,\sin\!\left(\frac{\eta\pi}{2}\right)\epsilon,
\quad
\epsilon \sim \mathcal{N}(0, I),
\label{eq:stochastic_update}
\end{equation}
with
\begin{equation}
\mu_\theta(z_t, t, \alpha)
= (1-\alpha)\,\hat{z}_0
+ \alpha\,\cos\!\left(\frac{\eta\pi}{2}\right)\hat{z}_1,
\label{eq:mu_def}
\end{equation}
where $\eta \in [0,1]$ controls exploration strength and $\sigma(t)$ decreases with $t$. When $\eta \to 0$, Eq.~\eqref{eq:stochastic_update} recovers the deterministic update in Eq.~\eqref{eq:deterministic_euler}, so AMP remains compatible with the original flow dynamics.

Intuitively, $(\hat{z}_0,\hat{z}_1)$ defines a local flow-consistent neighborhood, and the noise term explores randomized directions within this neighborhood rather than perturbing the trajectory arbitrarily. This property is important for aesthetic alignment: early steps with larger uncertainty can propose diverse motion hypotheses, whereas later steps become nearly deterministic and preserve identity, articulation, and temporal stability. Therefore, AMP provides a structured policy $p_\theta(z_{\alpha}\mid z_t)$ in latent space whose sampling behavior can be optimized by GRPO~\cite{grpo} toward human-aligned motion aesthetics in Stage~2. In our implementation, this aesthetic preference is instantiated by a reward ensemble composed of HPS-v2~\cite{hpsv2} and the visual-quality and motion-quality heads of VideoReward~\cite{videoreward}, while the detailed objective is deferred to Sec.~\ref{subsec:training_strategy}.

\noindent\textbf{Timestep-Truncated Aesthetic Alignment.}
We further observe that flow-based generation exhibits a coarse-to-fine structure: early denoising steps mainly determine global motion attributes such as head movement and overall expressiveness, whereas later steps refine high-frequency details that are critical for precise articulation and visual stability. Directly applying preference optimization to all timesteps would therefore entangle global aesthetic exploration with the local motion details already well learned in previous training stage. To preserve this strong articulation prior, we introduce timestep-truncated aesthetic alignment. Specifically, given a threshold $t_{\mathrm{thresh}} \in [0,1]$, we apply AMP and optimize it with GRPO only on the early stage $t \ge t_{\mathrm{thresh}}$, while the later stage $t < t_{\mathrm{thresh}}$ follows the deterministic flow update. This strategy allows the model to explore more human-preferred motion patterns at coarse scales without disturbing the fine-grained lip articulation already established by the flow-based motion generator and PKA.

\subsection{Training Strategy}
\label{subsec:training_strategy}

We train VibeAvatar in two stages. Stage~1 learns an articulation-aware flow prior under PKA conditioning. Stage~2 freezes PKA and aligns AMP with human aesthetic preference through GRPO, so that preference optimization improves motion aesthetics without disturbing the articulation prior learned in Stage~1.

\noindent\textbf{Mixed Preceding-frame Conditions.}
A key challenge is the distribution shift of the preceding motion condition $W_{\text{past}}$: supervised training naturally uses ground-truth history, whereas inference depends on model-generated history. To reduce this mismatch, we construct $W_{\text{past}}$ from a mixture of three sources:
\begin{equation}
W_{\text{past}} =
\begin{cases}
    \mathbf{0}, & \text{w.p. } p_0 \quad \text{(no history)}, \\
    W_{\text{GT}}, & \text{w.p. } p_{\text{gt}} \quad \text{(ground-truth history)}, \\
    W_{\text{model}}, & \text{w.p. } p_{\text{model}} \quad \text{(self-generated history)},
\end{cases}
\end{equation}
where $p_0 + p_{\text{gt}} + p_{\text{model}} = 1$. This is shared by both stages.

\begin{figure*}[ht!]
    \centering
    \includegraphics[width=0.9\linewidth]{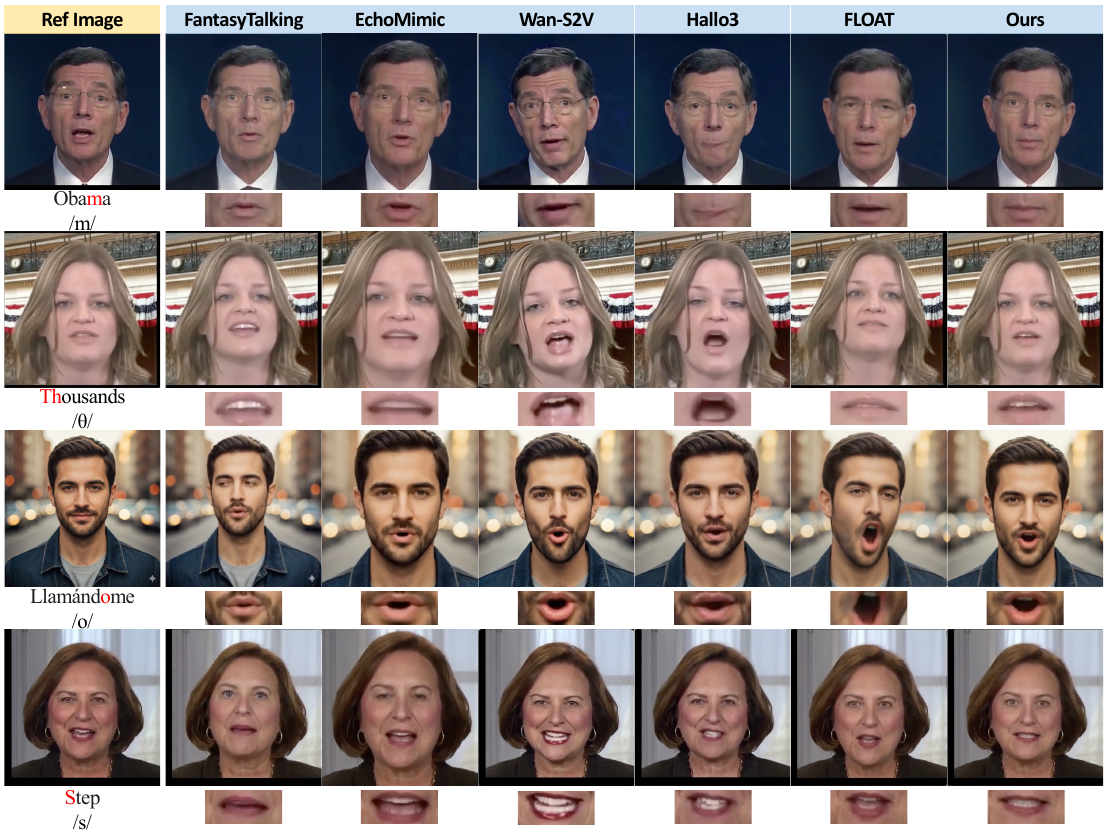}
    \caption{
    Lip-articulation comparison on challenging phonemes. VibeAvatar produces more accurate lip shapes.
    }
    \label{fig:qualitative_comparison_1}
    \vspace{-0.2cm}
\end{figure*}

\noindent\textbf{Stage~1: Flow-matching with PKA.}
In Stage~1, we sample a training clip $(S_{-L':L}, a_{-L':L})$, choose a reference frame $S_k$, and encode the target segment $\{S_0,\dots,S_L\}$ into motion latents $z_0 = w_{r\to S_{0:L}}$. We then draw $z_1 \sim \mathcal{N}(0,I)$ and $t \sim \mathrm{Uniform}(0,1)$, and form the interpolated latent
\begin{equation}
z_t = (1-t)\,z_0 + t\,z_1.
\end{equation}
The audio segment $a_{-L':L}$ is encoded by PKA to obtain conditioning $c$. Following Rectified Flow~\cite{rectflow}, we supervise the velocity field on the current clip by
\begin{equation}
\mathcal{L}_{\text{OT}}(\theta)
= \bigl\| v_{\theta}\bigl(z_t, t; W_{\text{past}}, w_{r\to S_k}, c\bigr)_{0:L}
      - (z_1 - z_0) \bigr\|.
\end{equation}
To encourage temporal consistency and effective use of history, we further regularize the prediction on the preceding window:
\begin{equation}
\mathcal{L}_{\text{con}}(\theta)
= \bigl\| v_{\theta}\bigl(z_t, t; W_{\text{past}}, w_{r\to S_k}, c\bigr)_{-L':0}
      - W_{\text{past}} \bigr\|.
\end{equation}
The Stage~1 objective is
\begin{equation}
\mathcal{L}_{\text{FM}}(\theta)
= \mathcal{L}_{\text{OT}}(\theta)
+ \lambda_{\text{con}} \mathcal{L}_{\text{con}}(\theta).
\end{equation}
This stage establishes a stable articulation-aware motion prior, which later serves as the base policy for AMP.

\noindent\textbf{Stage~2: GRPO on AMP.}
Starting from the pretrained parameters $\theta_0$, we freeze PKA and optimize AMP with GRPO over the truncated timestep range $t \ge t_{\mathrm{thresh}}$, while keeping the later stage deterministic. For each condition set, AMP generates a group of candidate motion trajectories over the early denoising stage. These candidates are decoded into videos and evaluated by a reward ensemble composed of HPS-v2~\cite{hpsv2} and the visual-quality and motion-quality heads of VideoReward~\cite{videoreward}. We aggregate these signals as
\begin{equation}
R^{(i)} 
= w_{\mathrm{ae}} R_{\mathrm{ae}}^{(i)}
+ w_{\mathrm{vq}} R_{\mathrm{vq}}^{(i)}
+ w_{\mathrm{mq}} R_{\mathrm{mq}}^{(i)},
\end{equation}
where $w_{\mathrm{ae}}, w_{\mathrm{vq}}, w_{\mathrm{mq}}$ are fixed coefficients. We then compute group-relative advantages and optimize the truncated AMP policy with the standard clipped GRPO objective. 
Overall, Stage~2 reshapes AMP to favor motion trajectories with higher human-aligned rewards at coarse motion scales, while preserving the fine-grained articulation and audio-motion alignment learned in Stage~1.

\section{Experiments}
\label{sec:experiment}

\subsection{Settings}

\begin{table*}[ht]
\caption{
   Quantitative comparisons. The best and second-best results are shown in bold and underlined, respectively.
}
\vspace{-0.2cm}

\centering
\small
\setlength{\tabcolsep}{3pt}
\renewcommand\arraystretch{1.1}
\resizebox{0.9\linewidth}{!}{
\begin{tabular}{l|ccccccc|ccccccc}
\toprule
\multirow{2}{*}{Method} & \multicolumn{7}{c|}{HDTF} & \multicolumn{7}{c}{RAVDESS} \\
\cmidrule(lr){2-8} \cmidrule(lr){9-15}
& FID$\downarrow$ & FVD$\downarrow$ & CSIM$\uparrow$ & Sync-C$\uparrow$ & Sync-D$\downarrow$ & VQA$\uparrow$ & ASE$\uparrow$
& FID$\downarrow$ & FVD$\downarrow$ & CSIM$\uparrow$ & Sync-C$\uparrow$ & Sync-D$\downarrow$ & VQA$\uparrow$ & ASE$\uparrow$ \\
\midrule
EchoMimic~\cite{echomimic}      & 40.776 & 352.494 & \underline{0.851} & 6.225 & 9.169 & 3.402 & 1.856 
               & 69.339 & 449.586 & 0.845 & 4.297 & 8.485 & 3.827 & 2.257 \\
FantasyTalking~\cite{fantasytalking} & 53.089 & 506.815 & 0.556 & 3.086 & 12.191 & 3.314 & 1.974 
               & 21.774 & 330.826 & \underline{0.852} & 4.009 & 9.322 & 3.748 & 2.330 \\
FLOAT~\cite{float}          & \underline{25.848} & \underline{220.927} & 0.842 & 7.345 & 8.172 & 3.616 & 
2.021 
               & \underline{18.754} & \underline{233.079} & 0.839 & 5.632 & \underline{7.771} & 3.882 & 2.358 \\
Wan-S2V~\cite{wan-s2v}        & 29.035 & 317.665 & 0.744 & \underline{7.669} & \underline{8.110} & \underline{3.629} & \underline{2.102} 
               & 18.824 & 233.341 & 0.828 & \underline{6.092} & 8.371 & \underline{4.016} & \underline{2.458} \\
Hallo3~\cite{hallo3}         & 26.277 & 253.995 & 0.729 & 7.648 & 8.571 & 3.497 & 2.024 
               & 24.418 & 237.575 & 0.823 & 5.421 & 8.046 & 3.679 & 2.103 \\
\midrule
VibeAvatar (Ours)     & \textbf{22.033} & \textbf{166.913} & \textbf{0.912} & \textbf{8.639} & \textbf{6.910} & \textbf{3.678} & \textbf{2.152} 
               & \textbf{16.282} & \textbf{220.975} & \textbf{0.866} & \textbf{6.134} & \textbf{7.439} & \textbf{4.121} & \textbf{2.577} \\
\bottomrule
\end{tabular}
}
\label{table:quantitative_comparisons}
\end{table*}

\begin{figure}[ht!]
    \centering
    \includegraphics[width=\linewidth]{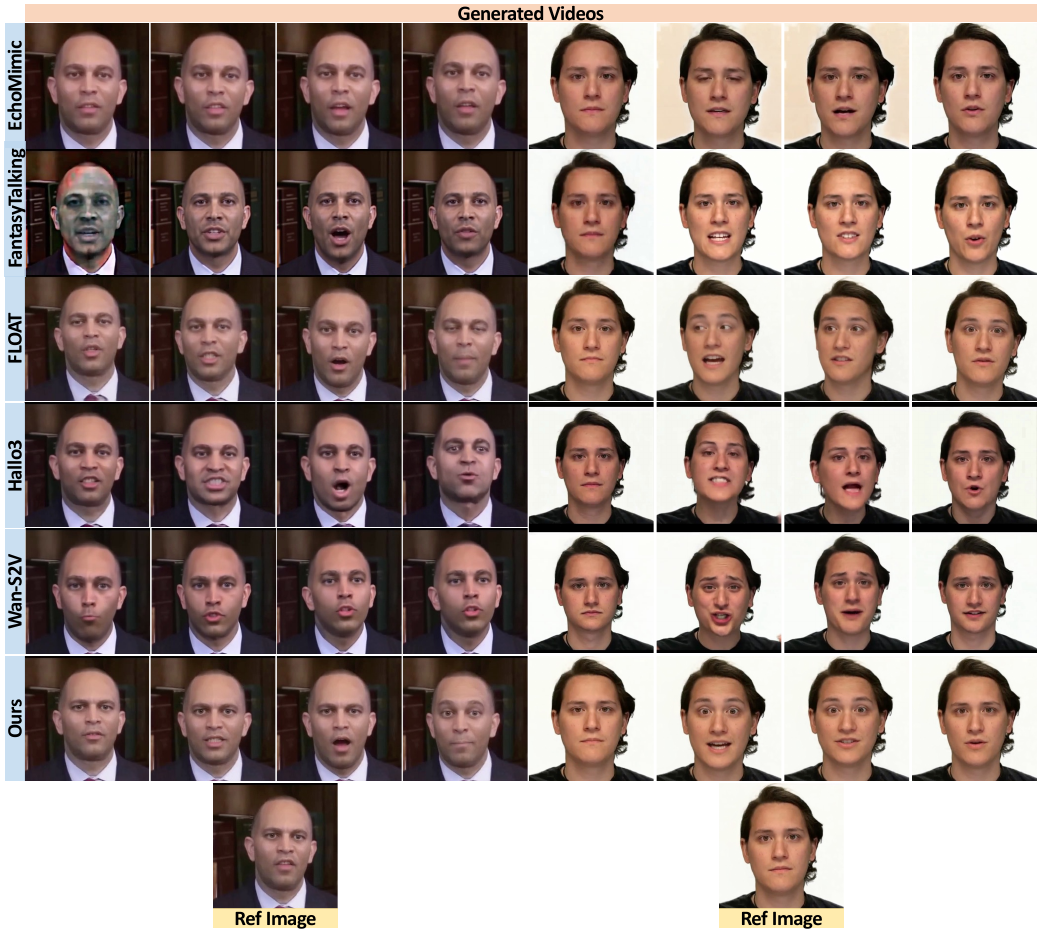}
    \caption{
   Multi-frame qualitative comparisons.
    }
    \label{fig:qualitative_comparison_2}
    \vspace{-0.5cm}
\end{figure}

\noindent\textbf{Datasets.}
We train VibeAvatar on HDTF~\cite{hdtf} and RAVDESS~\cite{ravdess}. We extract one subject per video, resample audio to 16 kHz, normalize to 25 FPS, and crop $512\times512$ face regions, discarding clips under 2s or severely corrupted.
To ensure fair evaluation, we build test sets with no overlap between training and testing speakers. For HDTF, we select $15$ videos from unseen speakers and uniformly crop $15$s segments, yielding a fixed $15\times15$s test set. For RAVDESS, we reserve $50$ full-length videos from $2$ actors as the test set, and use all remaining identities for training. The same test splits are used for all metrics, qualitative visualizations, and the user study.

\noindent\textbf{Evaluation Metrics.}
We evaluate performance using Sync-C~\cite{sync} and Sync-D~\cite{sync} for lip-sync accuracy, FID~\cite{cleanfid}, FVD~\cite{fvd} for visual and temporal quality, CSIM~\cite{deng2018arcface} for identity consistency and ASE and VQA from Q-Align~\cite{qalign} for aesthetic quality. A user study further assesses perceived realism and overall human preference.

\noindent\textbf{Comparison Methods.}
We compare VibeAvatar with state-of-the-art talking avatar methods, including FantasyTalking (MM'25)~\cite{fantasytalking},  EchoMimic (AAAI'25)~\cite{echomimic}, FLOAT  (ICCV'25)~\cite{float}, Wan-S2V (commercial model)~\cite{wan-s2v} and Hallo3  (CVPR'25) ~\cite{hallo3}. All methods are evaluated under recommended settings, using same reference image, driving audio, and held-out test videos for a fair comparison.

\subsection{State-of-the-Art Comparisons}

\noindent\textbf{Lip-articulation Comparison.}
\Cref{fig:qualitative_comparison_1} compares lip articulation on challenging phonemes that require accurate mouth closures and fine lip shapes. Existing methods often produce incomplete closures or ambiguous lip contours, which weakens articulation clarity and makes the speech look mumbled. In contrast, VibeAvatar better captures phonetic mouth configurations, producing clearer plosives and sharper lip transitions. This visual evidence is consistent with the stronger Sync-C and Sync-D results in \Cref{table:quantitative_comparisons}.

\begin{figure}[ht!]
    \centering
    \includegraphics[width=0.85\linewidth]{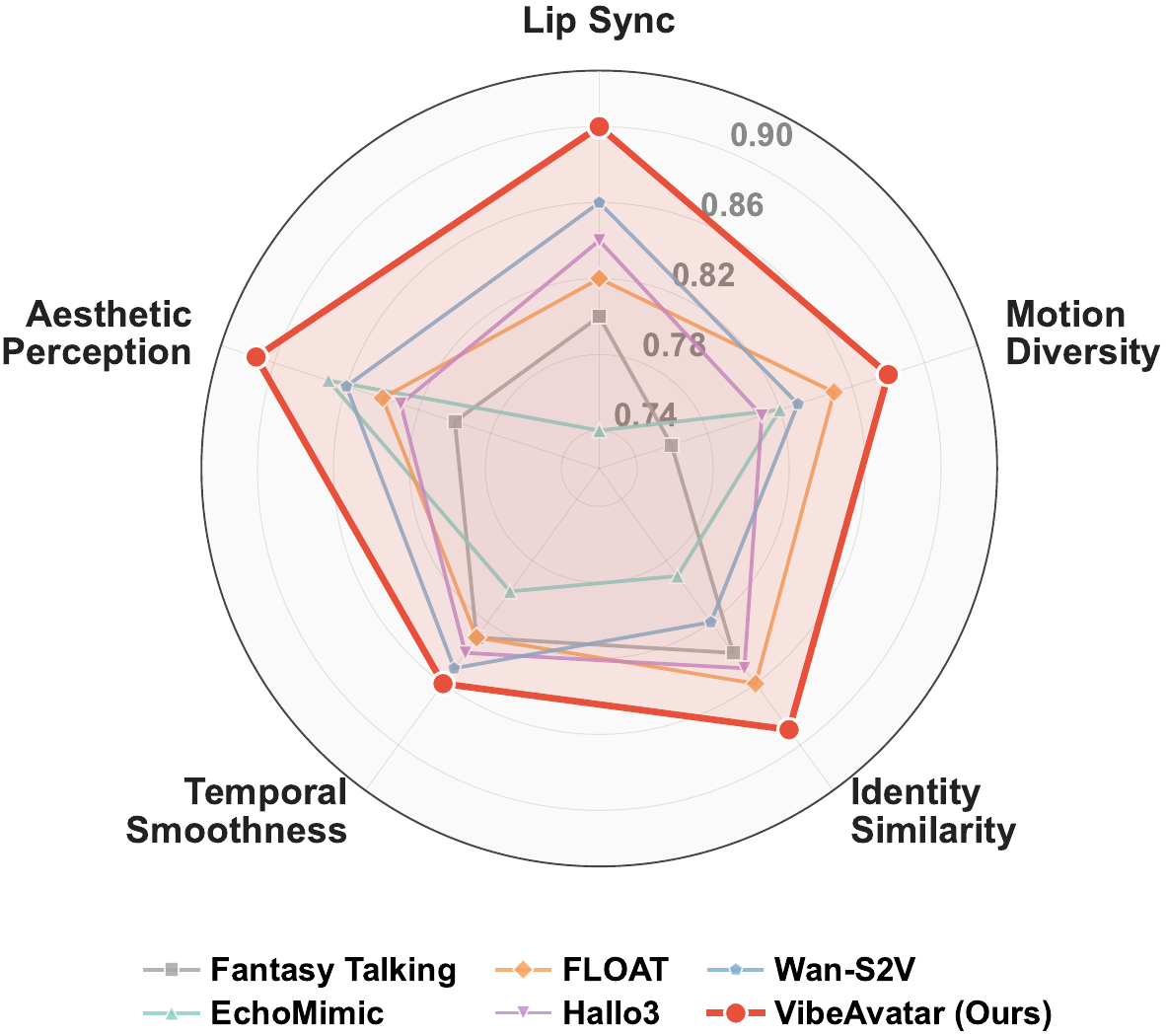}
    \caption{
    User study across five perceptual dimensions.
    }
    \label{fig:user_study}
    \vspace{-0.6cm}
\end{figure}

\noindent\textbf{Multi-frame Qualitative Comparison.}
\Cref{fig:qualitative_comparison_2} compares multiple frames from generated videos on HDTF and RAVDESS. Existing methods still show visible trade-offs among appearance sharpness, identity preservation, and motion coherence. EchoMimic tends to over-smooth facial details, FantasyTalking often introduces texture flickering and color artifacts, and FLOAT may drift in lip shape or identity under expressive motion. Hallo3 and Wan-S2V are visually stronger than these earlier baselines, but they still show noticeable issues in multi-frame consistency: Hallo3 occasionally produces exaggerated facial deformation around the mouth and cheeks, while Wan-S2V tends to yield less coherent expression transitions across frames. By contrast, VibeAvatar preserves cleaner facial details, more stable identity, and more coherent facial dynamics throughout the sequence, leading to stronger multi-frame visual quality.

\noindent\textbf{Quantitative Comparison.}
\Cref{table:quantitative_comparisons} reports comprehensive quantitative results on HDTF and RAVDESS. VibeAvatar achieves the best results on all reported metrics across both datasets, including Sync-C/Sync-D, FID/FVD, CSIM, VQA, and ASE. These results show stronger articulation accuracy, visual quality, identity preservation, and aesthetic quality than prior methods.

\begin{table*}[ht!]
\caption{
    Ablation results on HDTF and RAVDESS. The best and second-best results are shown in bold and underlined, respectively.
}
\vspace{-0.2cm}
\centering
\small
\setlength{\tabcolsep}{3pt}
\renewcommand\arraystretch{1.1}
\resizebox{0.9\linewidth}{!}{
\begin{tabular}{cc|ccccccc|ccccccc}
\toprule
\multirow{2}{*}{PKA} & \multirow{2}{*}{AMP} & \multicolumn{7}{c|}{HDTF} & \multicolumn{7}{c}{RAVDESS} \\
\cmidrule(lr){3-9} \cmidrule(lr){10-16}
& & FID$\downarrow$ & FVD$\downarrow$ & CSIM$\uparrow$ & Sync-C$\uparrow$ & Sync-D$\downarrow$ & VQA$\uparrow$ & ASE$\uparrow$
& FID$\downarrow$ & FVD$\downarrow$ & CSIM$\uparrow$ & Sync-C$\uparrow$ & Sync-D$\downarrow$ & VQA$\uparrow$ & ASE$\uparrow$ \\
\midrule
$\times$ & $\times$ 
         & \underline{26.135} & 215.733 & 0.829 & 7.523 & 8.232 & 3.583 & \underline{2.004} 
         & 19.480 & \underline{240.107} & \underline{0.850} & 5.890 & 7.925 & 3.887 & 2.379 \\
\checkmark & $\times$ 
         & 28.236 & \underline{190.595} & \underline{0.885} & \underline{8.615} & \underline{6.948} & \underline{3.595} & 1.921 
         & \underline{17.679} & 251.314 & 0.829 & \underline{6.042} & \underline{7.608} & \underline{3.892} & \underline{2.417} \\
\checkmark & \checkmark 
         & \textbf{22.033} & \textbf{166.913} & \textbf{0.912} & \textbf{8.639} & \textbf{6.910} & \textbf{3.678} & \textbf{2.152} 
         & \textbf{16.282} & \textbf{220.975} & \textbf{0.866} & \textbf{6.134} & \textbf{7.439} & \textbf{4.121} & \textbf{2.577} \\
\bottomrule
\end{tabular}
}
\label{table:ablation_core}
\vspace{-0.1cm}
\end{table*}

\noindent\textbf{User Study.} We further conduct a user study across five dimensions: Lip Sync, Motion Diversity, Identity Similarity, Temporal Smoothness, and Aesthetic Perception. Participants compare randomly ordered videos from all methods, and the normalized mean scores are summarized in \Cref{fig:user_study}. VibeAvatar receives the highest ratings in lip sync, aesthetic perception, and motion diversity, while remaining competitive in identity similarity and temporal smoothness. These results further suggest that our method better matches human perception in both realism and expressiveness.

\subsection{Ablation Study}

\begin{figure}[t!]
    \centering
    \includegraphics[width=0.9\linewidth]{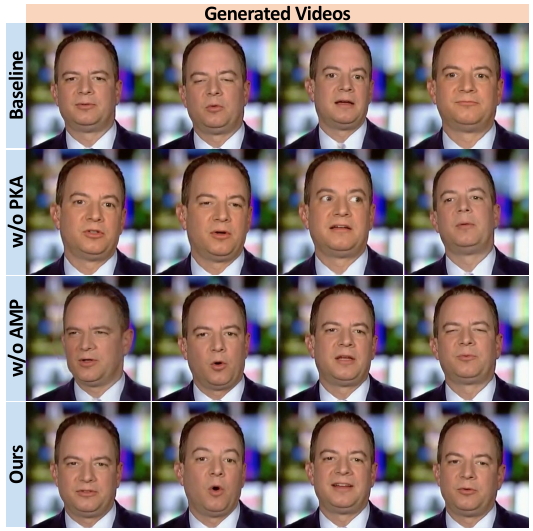}
    \vspace{-0.2cm}
    \caption{
    Qualitative ablation on the core components.
    }
    \label{fig:qualitative_ablation_core}
    \vspace{-0.4cm}
\end{figure}

We conduct ablations on the two core components of VibeAvatar, i.e., PKA and AMP. We further compare different audio context modeling strategies inside PKA.
In \Cref{table:ablation_core}, \emph{baseline} denotes the flow-based motion generator trained without PKA or AMP. Corresponding qualitative results are shown in \Cref{fig:qualitative_ablation_core}.

\noindent\textbf{Effectiveness of PKA.}
Adding PKA to the baseline mainly improves articulation accuracy and semantic alignment, which is exactly the behavior it is designed to target. The clearest gains appear on the synchronization metrics, showing that explicitly adapting generic speech features into phonetic-kinematic conditions is more effective than directly feeding recognition-oriented audio features into the generator. Although PKA alone does not improve every perceptual metric, it establishes a much stronger articulation prior, which later becomes the basis for the full model.

\noindent\textbf{Effectiveness of AMP.}
Adding AMP on top of PKA improves overall perceptual quality while preserving the articulation gains brought by PKA. Compared with the PKA-only model, AMP consistently improves fidelity, identity, and aesthetic metrics, while keeping synchronization strong. This pattern is important: AMP does not merely make videos look better in isolation, but improves human-preferred motion quality without sacrificing lip accuracy. In other words, PKA and AMP play different roles, with PKA building a strong articulation prior and AMP refining motion dynamics toward more realistic and aesthetically aligned behavior.

\noindent\textbf{Effectiveness of Audio Context Modeling Strategies.}
\Cref{tab:audio_modeling} compares audio context modeling strategies inside PKA. Causal attention already improves over the baseline, confirming that audio conditioning should respect the temporal structure of motion generation. Full attention and fixed chunk size bring further gains by using broader context, but they also expose the limitation of two naive extremes: full attention ignores causality, while fixed chunks use broader context without adapting receptive fields across layers. Our hierarchical shifted strategy performs best overall, suggesting that accurate lip articulation requires both local bidirectional interaction and progressively expanded phonetic context. This result provides direct evidence for the core design choice in PKA.

\begin{table}[t]
    \centering
    \caption{Ablation on audio context modeling strategies.}
    \vspace{-0.2cm}
    \label{tab:audio_modeling}
    \resizebox{0.9\linewidth}{!}{%
    \begin{tabular}{c|cc|cc}
        \toprule
        \multirow{2}{*}{Strategy} & \multicolumn{2}{c|}{HDTF} & \multicolumn{2}{c}{RAVDESS} \\
        \cmidrule(lr){2-3} \cmidrule(lr){4-5}
        & Sync-C $\uparrow$ & Sync-D $\downarrow$ & Sync-C $\uparrow$ & Sync-D $\downarrow$ \\
        \midrule
        baseline         & 7.523 & 8.232 & 5.890 & 7.925 \\
        causal           & 8.295 & 7.525 & 5.920 & 7.727 \\
        full attention   & 8.346 & \underline{7.328} & 5.991 & 7.686 \\
        fixed chunk & \underline{8.428} & 7.454 & \underline{6.009} & \underline{7.648} \\
        hierarchical shifted & \textbf{8.639} & \textbf{6.910} & \textbf{6.134} & \textbf{7.439} \\
        \bottomrule
    \end{tabular}%
    }
\vspace{-0.1cm}
\end{table}

\begin{table}[t]
    \centering
    \caption{Efficiency comparison on a 10-second 512px video.}
    \vspace{-0.2cm}
    \label{tab:efficiency}
    \resizebox{0.9\linewidth}{!}{%
    \begin{tabular}{lccc}
        \toprule
        Method & Params & Inference Time & VRAM \\
        \midrule
        EchoMimic~\cite{echomimic} & 3B & $\sim$8 min & \underline{$\sim$7 GB} \\
        FantasyTalking~\cite{fantasytalking} & 14B & $\sim$80 min & $\sim$51 GB \\
        Wan-S2V~\cite{wan-s2v} & 14B & $\sim$25 min & $\sim$57 GB \\
        Hallo3~\cite{hallo3} & 10B & $\sim$50 min & $\sim$70 GB \\
        FLOAT~\cite{float} & \textbf{0.6B} & \textbf{$<$10 s} & \underline{$\sim$7 GB} \\
        VibeAvatar (Ours) & \underline{0.7B} & \textbf{$<$10 s} & \textbf{$\sim$3 GB} \\
        \bottomrule
    \end{tabular}%
    }
\vspace{-0.2cm}
\end{table}

\subsection{Efficiency Analysis}

\Cref{tab:efficiency} reports the computational cost of generating a 10-second 512px video. VibeAvatar finishes inference in under 10 seconds and remains much faster than larger diffusion-based or commercial baselines that require minutes of generation. It is also lightweight in resource usage, with only $0.7$B parameters and about $3$ GB of VRAM, making it substantially more efficient than recent large models. These results show that VibeAvatar offers a favorable balance between generation quality and deployment cost.

\section{Conclusion}
\label{sec:conclusion}

We present VibeAvatar, an efficient multi-modal talking avatar framework built on the insight that phonetic accuracy and motion aesthetics arise from different sources and should be addressed at complementary stages. A Phonetic Kinematics Adapter (PKA) converts speech features into phonetic-kinematic conditions for accurate lip articulation, while an Aesthetic Motion Policy (AMP) aligns motion dynamics with human aesthetics through GRPO-based post-training without degrading the established articulation prior. Operating in a compact 1D warp-based latent motion space, VibeAvatar achieves state-of-the-art results on both objective metrics and user studies while maintaining practical efficiency.

\begin{acks}
This work is supported by the Fundamental Research Funds for the Central Universities, Peking University.
\end{acks}

\bibliographystyle{ACM-Reference-Format}
\balance
\bibliography{sigconf}

\end{document}